# A Dual-Polarization Information Guided Network for SAR Ship Classification

Tianwen Zhang, and Xiaoling Zhang

*Abstract*—How to fully utilize polarization to enhance synthetic aperture radar (SAR) ship classification remains an unresolved issue. Thus, we propose a dual-polarization information guided network (DPIG-Net) to solve it. DPIG-Net utilizes available dual-polarization information from Sentinel-1 SAR satellite to guide feature extraction and feature fusion. We first design a novel polarization channel cross-attention framework (PCCAF) for fine feature extraction. We establish a novel dilated residual dense learning framework (DRDLF) for fine feature fusion. Results on the open OpenSARShip dataset indicate DPIG-Net's state-of-the-art classification accuracy, compared with the other eleven competitive models. DPIG-Net can promote an effective and sufficient utilization of SAR polarization data in the future, of great value.

## I. INTRODUCTION

SHIP classification plays an important role in ocean surveillance. It can offer rich traffic information, helpful for trade developments. Synthetic aperture radar (SAR) is an active microwave sensor. Its operation is often unhindered by both light and weather, so it is very suitable for marine ship classification. Nowadays, SAR ship classification is receiving much attention.

Similar to SAR automatic target recognition (ATR) methods [1]–[5] designed for vehicle targets, traditional ship classification methods [6]–[11] focus on feature extraction based on experience, but the manual process is labor- and time-consuming. Modern deep learning methods [12]–[21] are receiving more attention. For example, Hou *et al.* [12] designed a simple convolutional neural network to classify ships in Gaofen-3 images. Huang *et al.* [13] proposed a group squeeze excitation sparsely connected convolutional network to extract robust ship features. Wang *et al.* [14] studied transfer learning to solve few-shot ship classification problem. Wang *et al.* [15] proposed a semi-supervised learning method via self-consistent augmentation to boost classification accuracy. He *et al.* [16] designed a densely connected triplet CNNs and integrated Fisher discrimination regularized metric learning for ship classification in medium-resolution SAR images. Zhang *et al.* [17] fused HOG features into CNNs to reduce model risk. However, these works did not consider ship polarization information, resulting in limited performance, especially for low-resolution SAR images.

Several works [18]–[21] tried to utilize polarization for better classification performance. For example, Zeng *et al.* [18] proposed a loss function for better dual-polarization feature training, but their network ignored feature interaction, which might lead to local optimization. Zhang *et al.* [19] designed a squeeze-and-excitation Laplacian pyramid network for multi-resolution feature extraction, but their network did not highlight more salient features, causing limited accuracy gains. Xiong *et al.* [20] established a mini hourglass region extraction network for dual-channel feature fusion, but they did not consider channel correlation, resulting in insufficient utilization of polarization information. Zhang *et al.* [21] established a polarization fusion and geometric feature embedding network to increase feature richness, but their network treated each polarization branch equally, resulting in difficult training and incomplete feature extraction.

From above, it remains a challenging and unresolved issue to make full use of polarization information to further boost SAR ship classification performance. Previous works [6]–[21] have not provided a simple and effective implementation way so far. Therefore, we propose a dual-polarization information guided network (DPIG-Net) to address it. DPIG-Net utilizes available dual-polarization information from Sentinel-1 satellite to guide SAR ship classification from two aspects — feature extraction and feature fusion. In the feature extraction process, we first design a novel polarization channel cross-attention framework (PCCAF) to model feature correlation, which can extract more representative features. In the feature fusion process, we design a novel dilated residual dense learning framework (DRDLF) to refine features, which can enable better feature fusion benefits. Results on the open three- and six-category OpenSARShip datasets [22] reveal the state-of-the-art classification accuracy of DPIG-Net, compared with the other eleven competitive models.

The main contributions of this paper are as follows.

1) DPIG-Net is proposed for the sufficient polarization utilization to boost classification accuracy. It is a brand-new architecture to achieve dual-polarization SAR ship classification. Compared with the other state-of-the-art methods, DPIG-Net can make more full use of ship polarization information, and has the potential to implicitly mine useful dual-polarization feature patterns for better classification accuracy.
2) PCCAF is proposed for the representative polarization feature extraction. It is a brand-new framework for dual-polarization feature extraction. Compared with the other state-of-the-art methods, PCCAF can model the correlation between different polarization channels by the proposed cross-attention subnetwork so as to serve for better feature extraction.
3) DRDLF is proposed for the refined polarization feature fusion. It is a brand-new framework to achieve dual-polarization feature fusion. Compared with the other state-of-the-art methods, DRDLF can maintain a large receptive field in network depth; its idea of feature reuse is conducive to the deep supervision of feature learning, reducing overfitting risk.
4) For the community of SAR ship detection and classification, we provide an idea of using polarization information to guide the intelligent interpretation of SAR images, and we contribute a network framework (PCCAF-DRDLF) which makes it possible to make full use of dual-polarization information.

The rest is organized as follows. Sec. II introduces DPIG-Net. Sec. III introduces experiments. Results are shown in Sec. IV. Discussions are described in Sec. V. Sec. VI sums up this letter.

This work was supported by the National Natural Science Foundation of China under Grant 61571099. *(Corresponding author: Xiaoling Zhang.)*

The authors are with the School of Information and Communication Engineering, University of Electronic Science and Technology of China, Chengdu 611731, China (e-mail: twzhang@std.uestc.edu.cn, xlzhang@uestc.edu.cn)

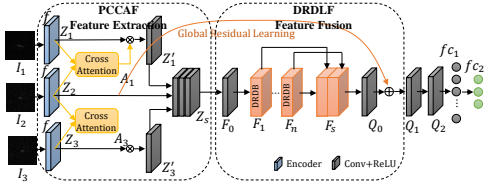

Fig. 1. Network architecture of DPIG-Net. PCCAF denotes the polarization channel cross-attention framework. DRDLF denotes the dilated residual dense learning framework. DRDB denotes the dilated residual dense block.

## II. DPIG-Net

Fig. 1 shows the network architecture of DPIG-Net. It is similar to [24], but it is closely related to ship polarization. The data used in this work is the open OpenSARShip dataset, whose images samples are collected from Sentinel-1 SAR satellite. Sentinel-1 works in dual-polarization mode, i.e., vertical-vertical (VV) and vertical-horizontal (VH). The offered data is denoted by $S_{VV}$ and $S_{VH}$ which are plural. Since $S_{VV}$ has higher scattering energy of ships [23], it is selected as the middle main branch, i.e. $I_2 = |S_{VV}|$. $S_{VH}$ reflects less scattering energy of ships than $S_{VV}$ [23], so it is selected as the upper branch, i.e. $I_1 = |S_{VH}|$. See [23] for more descriptions. Moreover, the down branch in PCCAF is used to measure polarization channel difference for more comprehensive description of ship characteristics, and its input is

$$I_3 = |S_{VV} \cdot S_{VH}^*| \quad (1)$$

where * denotes a complex conjugate operation. Thus, $S_{VV}$ and $S_{VH}$ used in our work must be complex data, rather than the previous commonly-used amplitude-based real data. To the best of our knowledge, OpenSARShip might be the only data that can meet this condition. Noted that FUSAR-Ship [12] only offered amplitude real data, so $I_3$ cannot be obtained by Eq. (1). Moreover, images in FUSAR-Ship are not paired in the form of VV-VH or HH-HV, which makes our network unable to apply to it.

Especially, our current work only considers dual-polarization case due to the limitation of available data. If full-polarization data is available in the future, one can expand DPIG-Net into four parallel branches to receive four different polarization inputs (or more branches for the cross-channel modeling).

PCCAF receives three types of data ($I_1$, $I_2$, and $I_3$) for feature extraction. Its output is $Z_S$ which contains high-level semantic features of three types of data. DRDLF receives $Z_S$ for feature fusion using some dilated residual dense blocks and a global residual learning from the main branch $I_2$. Finally, 2D feature maps are flattened into 1D feature vector to transmit into a fully-connected ($fc1$) layer. The terminal $fc2$ is responsible for the category prediction with a soft-max function.

DPIG-Net shows a tendency of feature aggregation from input three branches to the terminal feature integration. Previous works usually only adopted $I_2$ to predict ship categories, i.e., the middle main branch of PCCAF. Differently, we make full use of polarization information ($I_1$ and $I_3$) to guide the classification prediction of $I_2$. We call the above paradigm the dual-polarization information guided SAR ship classification.

### A. Polarization Channel Cross-Attention Framework (PCCAF)

First, PCCAF establishes a simple encoder $f$ to preliminarily extract features from three types of data. The encoder structure is shown in TABLE I. The encoder $f$ uses standard convs to extract features, batch normalization (BN) [25] to ensure training, and ReLU to activate neurons. The max-pooling operation is to reduce the size of feature maps. With network deepening, the channel width increases by a multiple of 2. Our feature encoder $f$ only has four stages, rather than usual five stages [17]. This is to avoid the loss of spatial features due to the small size of SAR ships. Their outputs are denoted by $Z_1$, $Z_2$, and $Z_3$ for the subsequent processing. Note that more advanced encoder might bring better performance, but it is not the scope of this letter.

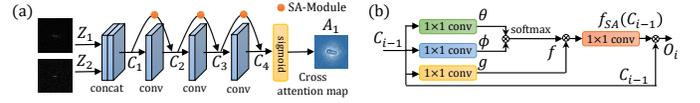

Fig. 2. Implementation of the cross-attention subnetwork.

TABLE I
ENCODER STRUCTURE IN PCCAF.

| Stage | Layer | Input Shape | Output Shape | Kernel@Stride |
|---|---|---|---|---|
| $S_1$ | Conv + BN + ReLU | 224×224×1 | 224×224×8 | 3×3×8@1 |
|       | Max-pooling | 224×224×8 | 128×128×8 | @2 |
| $S_2$ | Conv + BN + ReLU | 128×128×8 | 128×128×16 | 3×3×16@1 |
|       | Max-pooling | 128×128×16 | 64×64×16 | @2 |
| $S_3$ | Conv + BN + ReLU | 64×64×16 | 64×64×32 | 3×3×32@1 |
|       | Max-pooling | 64×64×32 | 32×32×32 | @2 |
| $S_4$ | Conv + BN + ReLU | 32×32×32 | 32×32×64 | 3×3×64@1 |
|       | Max-pooling | 32×32×64 | 16×16×64 | @2 |

We design a cross-attention subnetwork to model the correlation between different polarization branches. The design concept of the cross-attention subnetwork is that the middle main branch generates referenced feature maps to guide the other two auxiliary branches. Most existing attention networks [27], [29] merely refine their own feature maps in the uncrossed mode, which cannot solve multi-branch dual-polarization-guided case. That is, their module input has only one entry, but our proposed cross-attention subnetwork is specially designed for dual-polarization ship mission, i.e., our module input has two entries. The cross-attention subnetwork can be summarized as

$$A_i = \alpha_i (Z_i, Z_r) \quad (2)$$

where $Z_r$ denotes the referenced feature maps (in this paper, $Z_r = Z_2$, i.e., the main VV branch), $Z_i$ denotes the feature maps to be corrected (in this paper, $Z_i$ means the VH branch $Z_1$ or the polarization difference branch $Z_3$), $\alpha_i$ denotes the learned mapping, and $A_i$ denotes the cross-attention map.

Fig. 2(a) shows its network implementation. We take $Z_1$ and $Z_2$ as an example to introduce; the same to $Z_3$ and $Z_2$. We first concatenate the two input feature maps directly, and then, three convs with a skip connection are employed to learn inputs' interrelation. Finally, the learning knowledge will be activated by a sigmoid to obtain the final cross-attention maps $A_1$.

Furthermore, for better skip connection fusion between shallow low-level features and deep high-level features, we design a self-attention module (SA-Module) to refine previous features. SA-Module's motivation is also related to SAR image characteristics, e.g. speckle noises and sea clutters. It can relieve their interferences to enhance ships' saliency, as shown in Fig. 2(a). SA-Module can highlight more important global information in space, suppress low-value information, promoting network information flow. Ablation studies in Sec. V-A indicate that it can offer a ~2% accuracy improvement on the six-category task. It generates a self-attention map to modify input and then the result is added to the raw conv branch. The above is described as

$$C_i = C_{i-1} \cdot f_{SA}(C_{i-1}) + f_{3\times 3}(C_{i-1}) \quad (3)$$

where $C_i$ denotes the $i$-th conv feature map, $f_{SA}$ denotes the SA-

Module operation, $f_{3\times3}$ denotes the 3×3 conv. Fig. 2(b) shows the implementation process of SA-Module. Spatial features of the *i*-position are denoted by $\phi$ by a 1×1 conv learning. Spatial features of the *j*-position are denoted by $\theta$ by another one 1×1 conv learning. The relationship between *i*-position and *j*-position is denoted by *f* which is obtained by an adaptive learning between $\phi$ and $\theta$, where the normalization process is equivalent to a soft-max function. The representation of the input at *j*-position is denoted by *g* learned by another one 1×1 conv. The response at *i*-position is obtained by a matrix element-wise multiplication between input $C_{i-1}$ and self-attention map $f_{SA}(C_{i-1})$.

The final resulting cross-attention map is acted on the other two branches by matrix element-multiplication to obtain refined polarization-guided features, i.e.,

$$Z'_i = Z_i \otimes A_i \qquad (4)$$

where $Z_i'$ denotes polarization-guided features that will be used to guide the main polarization branch.

Finally, the output of the main polarization branch is the concatenation of three types of features, i.e.,

$$Z_S = \text{Concat}\left(Z'_1, Z_2, Z'_3\right) \qquad (5)$$

where $Z_S$ denotes the output of PCCAF. We find that the feature concatenation performs better than feature adding, because the former can avoid the resistance effects between different polarization features with our subsequent feature fusion operations.

### B. Dilated Residual Dense Learning Framework (DRDLF)

DRDLF uses some dilated residual dense blocks (DRDBs) to fuse the extracted polarization features coming from the previous PCCAF stage. The input of DRDLF is $Z_S$ which is associated with dual-polarization information using the concatenation operation of Eq. (5) where $Z_1'$ denotes the feature maps of $I_1$ VH information, $Z_2$ denotes that of $I_2$ VV information, and $Z_3'$ denotes that of VV-VH correlation information. $Z_S$ is refined by a 3×3 conv for feature concentration and channel dimensionality reduction. The result is denoted by $F_0$. Then, several DRDBs are used for feature aggregation. DRDB is motivated by RDB [26] designed for image super-resolution tasks. However, there are many speckle noises around SAR ship images, so we insert a dilated rate 2 to the standard conv for larger receptive fields.

Fig. 3 shows DRDB's implementation. Its input is the previous output $F_i$, and its output is denoted by $F_{i+1}$. DRDB contains three 2-dilated 3×3 conv layers whose results are denoted by $D_1$, $D_2$, and $D_3$. They are concatenated directly as $D_S$. For the residual learning in the entire DRDB, a 1×1 conv is used for channel reduction. Finally, the sum between $F_i$ and $D_S$ is its output. In DRDLF, we arrange *n* DRDBs for feature fusion where *n* is set to the optimal value 3 empirically. The results of *n* DRDBs from $F_1$ to $F_n$ are concatenated and then processed by a 1×1 conv for overall channel reduction. The result is denoted by $Q_0$.

We observe that after a series of DRDB processing, the details of the main VV branch might be gradually diluted, causing unstable training and deteriorating performance. Thus, we propose a global residual learning to solve this problem. As in Fig. 1, the global residual learning connects PCCAF and DRDLF that can maintain the dominant position of the main branch and make the other two branches smoothly play an auxiliary guiding role. This is an important design idea of our dual-polarization guided network. The global residual learning is described by

$$Q_1 = Q_0 + Z_2 \qquad (6)$$

where $Q_1$ denotes the final output of DRDLF. From Fig. 1, we set another two 3×3 convs to process $Q_1$ for more semantic features $Q_2$, helpful for balancing spatial and semantic information.

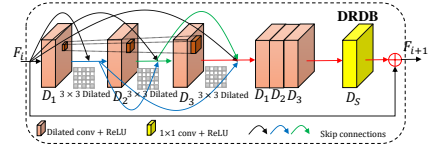

Fig. 3. Implementation of the dilated residual dense block (DRDB).

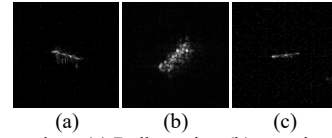

| TABLE II THREE-CATEGORY DATA. | | | TABLE III SIX-CATEGORY DATA. | | |
|---|---|---|---|---|---|
| Category | Training | Test | Category | Training | Test |
| Bulk carrier | 169 | 164 | Bulk carrier | 100 | 233 |
| Container ship | 169 | 404 | Cargo | 100 | 571 |
| Tanker | 169 | 73 | Container ship | 100 | 473 |
| | | | Fishing | 100 | 25 |
| | | | General cargo | 100 | 42 |
| | | | Tanker | 100 | 142 |

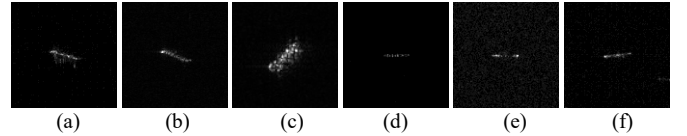

Fig. 4. Three-category data. (a) Bulk carrier; (b) container ship; (c) tanker.

Fig. 5. Six-category data. (a) Bulk carrier; (b) cargo; (c) container ship; (d) fishi general cargo; (f) tanker.

To sum up, combined with the above designed PCCAF and DRDLF, our proposed DPIG-Net can make full use of the polarization information ignored in previous works. The other two types of polarization data are well refined to assist the feature extraction and feature fusion of the main branch. Finally, an effective dual-polarization information guided SAR ship classification paradigm is realized. DPIG-Net successfully handles the problems of how to conduct polarization guidance and how to carry out more effective polarization guidance, of great value.

## III. EXPERIMENTS

### A. Dataset

The open OpenSARShip dataset [22] is used to evaluate the effectiveness of DPIG-Net. It offers VV-VH dual-polarization SAR ship data from Sentinel-1. The raw data is the single look complex (SLC) type. Same as [19], its two subsets are used for experiments, i.e., a three-category subset and a six-category one. As mentioned before, OpenSARShip is the only one dataset that can satisfy our experimental requirements, i.e., paired dual-polarization complex data with corresponding ground truth labels. TABLE II and TABLE III show more data descriptions. Fig. 4 and Fig. 5 show some samples of different ship categories.

### B. Training Details

We train DPIG-Net by 100 epochs from scratch using Adam with a learning rate of 0.0001. The network parameters are initialized by [30]. Samples are resized to 224×224 by bilinear interpolation. The batch size is set to 16. The multi-category cross entropy [15] serves as the loss function of networks. We reproduce other models basically consistent with their raw reports.

### C. Evaluation Criteria

TABLE IV
CLASSIFICATION PERFORMANCE OF DIFFERENT MODELS.

| Methods | Three-Category Acc (%) | Six-Category Acc (%) | Time (ms) |
|---|---|---|---|
| Hou *et al.* [12] | 67.41±1.13 | 47.44±2.01 | 4.30 |
| GSESCNN [13] | 74.98±1.46 | 54.78±2.08 | 4.28 |
| Wang *et al.* [14] | 69.27±0.27 | 48.43±3.71 | 4.33 |
| HOG-ShipCLSNet [17] | 75.44±2.68 | 54.93±2.61 | 4.52 |
| Zeng *et al.* [18] | 77.41±1.74 | 55.26±2.36 | 4.47 |
| SE-LPN-DPFF [19] | 79.25±0.83 | 56.66±1.54 | 5.05 |
| Mini Hourglass Net [20] | 75.44±2.68 | 54.93±2.61 | 4.52 |
| PFGFE-Net [21] | 79.84±0.53 | 56.83±2.68 | 4.95 |
| VGGNet-Grey [31] | 78.51±0.93 | 55.80±2.05 | 4.63 |
| GBCNN [32] | 78.84±0.26 | 56.48±1.94 | 4.85 |
| DenseNet-LRCS [33] | 78.00±0.00 | 56.29±0.00 | 5.36 |
| **DPIG-Net (Ours)** | **81.28±0.65** | **58.68±2.02** | 5.12 |

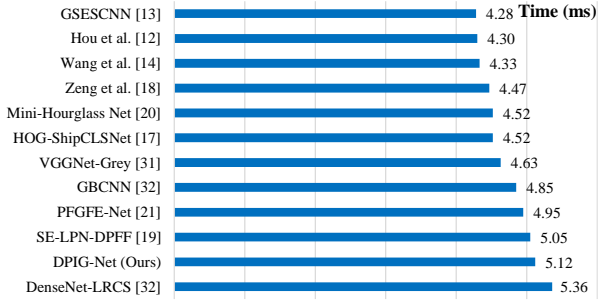

Fig. 6. Classification time comparison. Time is sorted from small to large.

TABLE V
CONFUSION MATRIX ON THE THREE-CATEGORY TASK.

| True \ Predicted | Bulk carrier | Container ship | Tanker |
|---|---|---|---|
| Bulk carrier | 125 | 21 | 8 |
| Container ship | 48 | 342 | 14 |
| Tanker | 11 | 8 | 54 |

TABLE VI
CONFUSION MATRIX ON THE SIX-CATEGORY TASK.

| True \ Predicted | Bulk carrier | Cargo | Container ship | Fishing | General cargo | Tanker |
|---|---|---|---|---|---|---|
| Bulk carrier | 143 | 23 | 43 | 0 | 22 | 2 |
| Cargo | 69 | 325 | 19 | 27 | 83 | 48 |
| Container ship | 67 | 16 | 359 | 0 | 29 | 2 |
| Fishing | 0 | 2 | 0 | 22 | 0 | 1 |
| General cargo | 6 | 20 | 2 | 0 | 9 | 5 |
| Tanker | 13 | 91 | 1 | 4 | 19 | 14 |

The accuracy (*Acc*) equals (*TP*+*TN*)/(*TP*+*TN*+*FP*+*FN*). *TP* denotes the true positives, *TN* denotes the true negatives, *FP* denotes the false positives, and *FN* denotes the false negatives.

IV. RESULTS

*A. Classification Performance*

*1) Accuracy.* TABLE IV is the quantitative evaluation of different models. The top-10 best results among 20 trainings are used to calculate the average and standard deviation except for DenseNet-LRCS [33]. DPIG-Net outperforms the other eleven comparative models obviously. The second-best model offers a 79.84% accuracy on the three-category task which is still lower than ours by 1.44%, and a 56.83% accuracy on the six-category task which is still lower than ours by 1.85%. This reveals the state-of-the-art classification performance of DPIG-Net. Note that such accuracy increment is already a huge progress in the SAR ship classification community. Compared with the other methods, DPIG-Net can make more full use of ship polarization information, and has the potential to implicitly mine useful dual polarization feature patterns for better classification accuracy.

*2) Computational Efficiency.* Fig. 6 shows the classification time comparison with different methods. DPIG-Net consumes more time (5.12ms) to classify ships than most other methods, but it is still faster than DenseNet-LRCS [33]. Furthermore, the speed gap between DPIG-Net and other methods is relatively small (within 1ms), so DPIG-Net might still meet practical applications. According to our theoretical statistics of network parameters, DPIG-Net has about 17,961,536 (~18M) parameters. This indicates that DPIG-Net might be a little heavy, which exactly leads to its more running time in our experiments as in Fig. 6. Thus, the speed optimization will be studied in the future.

*B. Confusion Matrix*

TABLE V–VI are the confusion matrix of DPIG-Net. DPIG-Net can identify most ships successfully, i.e., the diagonal value is greater than others at the same line in most cases.

V. DISCUSSIONS

*A. Discussion on PCCAF*

To confirm the effectiveness of PCCAF, we conduct some ablation studies on it, including the polarization-guided practice and the proposed cross-attention module. The results are shown in TABLE VII. From TABLE VII, the polarization-guided practice can offer obvious accuracy gains. Taking the six-category task as an example, $I_1$ (the VH polarization channel) boosts the accuracy by 1.47%, and $I_3$ (the polarization channel difference) boosts the accuracy by 2.67%. The combination of two inputs is better than the single alone; the combination of three inputs is better than the combination of two inputs. The above shows the effectiveness of polarization information. Moreover, the offered accuracy gain is greater than some previous works [19], [21]. This shows that PCCAF can make more full use of polarization information. Finally, the proposed cross-attention module can improve the classification accuracy further (a ~2% improvement on the six-category task), which is in line with the subjective analysis in Sec. II-A. This is because it can establish the correlation between channels to extract features with more mutual recognition. As a result, the information flow between channels is promoted for better feature extraction.

We discuss the effect of different inputs in the main branch on results as shown in TABLE VIII. The VV $I_2$ offers better results than others since it contains more ship scattering energy.

We conduct another one study to verify the advantage of feature concatenation over feature adding. Results are in TABLE IX. The former performs better than the latter, so features between different polarization channels should better not be added directly; otherwise it may cause feature resistance effects.

Finally, we perform experiments to confirm the effectiveness of SA-Module in the cross-attention subnetwork in TABLE X. SA-Module improves accuracy further since it can enable more prominent features for multi-stage residual fusion. Furthermore, SA-Module can ease negative effects of speckle noises and sea clutters of SAR characteristics, to enhance ships' saliency, as in Fig. 2(a). This is in line with experimental results in TABLE X.

TABLE VII
DISCUSSION RESULTS ON PCCAF.

| $I_2$ | Polarization-Guided $I_1$ | $I_3$ | Cross Attention | Three-Category Acc (%) | Six-Category Acc (%) |
|---|---|---|---|---|---|
| ✓ |  |  | -- | 78.45±1.78 | 52.69±2.98 |
|  | ✓ |  | -- | 77.86±1.94 | 50.82±1.75 |
|  |  | ✓ | -- | 75.28±1.50 | 50.32±2.03 |
| ✓ | ✓ |  | -- | 79.89±2.01 | 54.16±2.57 |
| ✓ | ✓ | ✓ | ✗ | 80.86±1.08 | 56.83±1.93 |
| ✓ | ✓ | ✓ | ✓ | **81.28±0.65** | **58.68±2.02** |

TABLE VIII
RESULTS OF DIFFERENT MAIN BRANCHES IN PCCAF.

| Main Branch | Three-Category Acc (%) | Six-Category Acc (%) |
|---|---|---|
| $I_1$ | 80.02±0.84 | 57.45±1.85 |
| $I_2$ | **81.28±0.65** | **58.68±2.02** |
| $I_3$ | 75.38±1.64 | 51.52±2.36 |

TABLE IX
RESULTS OF FEATURE CONCATENATION OR FEATURE ADDING IN PCCAF.

| Type | Three-Category Acc (%) | Six-Category Acc (%) |
|---|---|---|
| Feature Adding | 80.65±1.26 | 57.66±2.14 |
| **Feature Concatenation** | **81.28±0.65** | **58.68±2.02** |

TABLE X
RESULTS ON EFFECTIVENESS OF SA-MODULE.

| SA-Module | Three-Category Acc (%) | Six-Category Acc (%) |
|---|---|---|
| ✗ | 80.98±0.87 | 57.48±2.35 |
| ✓ | **81.28±0.65** | **58.68±2.02** |

TABLE XI
DISCUSSION RESULTS ON DRDLF.

| DRDB | Global Residual Learning | Three-Category Acc (%) | Six-Category Acc (%) |
|---|---|---|---|
| -- | -- | 79.44±0.82 | 55.38±1.98 |
| ✓ |   | 80.98±0.63 | 57.46±2.25 |
| ✓ | ✓ | **81.28±0.65** | **58.68±2.02** |

TABLE XII
RESULTS ON DIFFERENT NUMBERS OF DRDBS.

| Number | Three-Category Acc (%) | Six-Category Acc (%) |
|---|---|---|
| 1 | 80.69±0.48 | 57.05±2.26 |
| 2 | 80.99±0.32 | 57.87±2.18 |
| **3** | **81.28±0.65** | **58.68±2.02** |
| 4 | 81.02±0.17 | 58.27±3.01 |
| 5 | 80.78±0.84 | 58.02±3.18 |

*B. Discussion on DRDLF*

To verify the effectiveness of DRDLF, we conduct some ablation studies on it. The results are shown in TABLE XI. DRDB improves the accuracy by 1.54% on the three-category task and by 2.08% on the six-category task. It can learn context information more effectively for achieve more concentrated feature fusion effects. Furthermore, the global residual learning boosts the accuracy further, because it can restore original feature details from the main branch $I_2$ effectively, which avoids possible feature loss from multi conv and pooling operations.

We determine the number of DRDB empirically via experiments in TABLE XII. From TABLE XII, the accuracy increases first and then decreases as the increase of the number of DRDBs. One possible reason is that excessive DRDBs may lead to overfitting for its large number of network parameters. For this, we set the number of DRDB to the optimal value 3.

## VI. CONCLUSIONS

DPIG-Net is designed for dual-polarization guided SAR ship classification. PCCAF is designed for better dual-polarization feature extraction. DRDLF is designed for fine dual-polarization feature fusion. DPIG-Net utilized available dual-polarization information from Sentinel-1 to guide better ship classification. We perform extensive experiments on the public OpenSARShip dataset to confirm the effectiveness of DPIG-Net. Results reveal the state-of-the-art classification accuracy of DPIG-Net, compared with the other eleven competitive models.

*Limitations and Future Works.* The running speed of DPIG-Net is not attractive, so we will optimize its speed. Moreover, DPIG-Net merely considers polarization guidance at the feature map level, so other levels, e.g., decision-level, will be studied.

## REFERENCES

[1]. M. Amrani *et al.*, "SAR-oriented visual saliency model and directed acyclic graph support vector metric based target classification," *IEEE J. Sel. Top. Appl. Earth Obs. Remote Sens.*, vol. 11, no. 10, pp. 3794–3810, 2018.

[2]. M. Amrani *et al.*, "New SAR target recognition based on YOLO and very deep multi-canonical correlation analysis," *Int. J. Remote Sens.*, pp. 1, 2021.

[3]. A. Moussa and J. Feng, "Deep feature extraction and combination for synthetic aperture radar target classification," *J. Appl. Remote Sens.*, vol. 11, no. 4, pp. 1–18, 2017.

[4]. M. Amrani *et al.*, "An efficient feature selection for SAR target classification," in *Proc. Pacific Rim Conf. Multimedia.*, 2017, pp. 68–78.

[5]. A. Moussa *et al.*, "Bag-of-visual-words based feature extraction for SAR target classification," in *Proc. CDIP*, 2017, pp. 104201.

[6]. G. Margarit *et al.*, "On the usage of GRECOSAR, an orbital polarimetric SAR simulator of complex targets, to vessel classification studies," *IEEE Trans. Geosci. Remote. Sens.*, vol. 44, no. 12, pp. 3517–3526, 2006.

[7]. X. Xing, K. Ji, H. Zou, W. Chen, and J. Sun, "Ship classification in TerraSAR-X images with feature space based sparse representation," *IEEE Geosci. Remote Sens. Lett.*, vol. 10, no. 6, pp. 1562–1566, 2013.

[8]. H. Lang *et al.*, "Ship classification in sar image by joint feature and classifier selection," *IEEE Geosci. Remote Sens. Lett.*, vol.13, no.2, pp.212, 2016.

[9]. G. Margarit *et al.*, "Ship classification in single-pol SAR images based on fuzzy logic," *IEEE Trans. Geosci. Remote. Sens.*, vol. 49, pp. 3129, 2011.

[10]. M. Jiang, X. Yang, Z. Dong, S. Fang, and J. Meng, "Ship classification based on superstructure scattering features in SAR images," *IEEE Geosci. Remote Sens. Lett.*, vol. 13, no. 5, pp. 616–620, 2016.

[11]. Y. Xu *et al.*, "Discriminative adaptation regularization framework-based transfer learning for ship classification in SAR images," *IEEE Geosci. Remote Sens. Lett.*, vol. 16, no. 11, pp. 1786–1790, 2019.

[12]. X. Hou *et al.* "FUSAR-Ship: building a high-resolution SAR-AIS matchup dataset of Gaofen-3 for ship detection and recognition," *Sci. China Inf. Sci.*, vol. 63, no. 4, pp. 140303, 2020.

[13]. G. Huang, X. Liu, J. Hui, Z. Wang, and Z. Zhang, "A novel group squeeze excitation sparsely connected convolutional networks for SAR target classification," *Int. J. Remote Sens.*, vol. 40, no. 11, pp. 4346–4360, 2019.

[14]. Y. Wang *et al.* "Ship classification in high-resolution SAR images using deep learning of small datasets," *Sensors*, vol. 18, no. 9, pp. 2929, 2018.

[15]. C. Wang, J. Shi, Y. Zhou, X. Yang, Z. Zhou, S. Wei, and X. Zhang, "Semisupervised learning-based SAR ATR via self-consistent augmentation," *IEEE Trans. Geosci. Remote. Sens.*, vol. 59, no. 6, pp. 4862–4873, 2020.

[16]. J. He, Y. Wang, and H. Liu, "Ship classification in medium-resolution SAR Images via densely connected triplet CNNs integrating fisher discrimination regularized metric learning," *IEEE Trans. Geosci. Remote. Sens.*, vol. 59, no. 4, pp. 3022–3039, 2021.

[17]. T. Zhang *et al.*, "HOG-ShipCLSNet: A novel deep learning network with HOG feature fusion for SAR ship classification," *IEEE Trans. Geosci. Remote. Sens.*, vol. 60, no. 5210322, pp. 5210322, 2021.

[18]. L. Zeng *et al.*, "Dual-polarized SAR ship grained classification based on CNN with hybrid channel feature loss," *IEEE Geosci. Remote Sens. Lett.*, vol. 19, pp. 4011905, 2021.

[19]. T. Zhang, and X. Zhang, "Squeeze-and-excitation Laplacian pyramid network with dual-polarization feature fusion for ship classification in SAR images," *IEEE Geosci. Remote Sens. Lett.*, vol. 19, pp. 4019905, 2021.

[20]. G. Xiong, Y. Xi, D. Chen, and W. Yu, "Dual-polarization SAR ship target recognition based on mini hourglass region extraction and dual-channel efficient fusion network," *IEEE Access*, vol. 9, pp. 29078–29089, 2021.

[21]. T. Zhang *et al.*, "A polarization fusion network with geometric feature embedding for sar ship classification," *Pattern Recognit.*, vol. 123, pp. 8, 2021.

[22]. L. Huang, B. Liu, B. Li, W. Guo, W. Yu, Z. Zhang, and W. Yu, "OpenSARShip: A dataset dedicated to Sentinel-1 ship interpretation," *IEEE J. Sel. Top. Appl. Earth Obs. Remote Sens.*, vol. 11, no. 1, pp. 195–208, 2018.

[23]. R. Pelich *et al.*, "Large-scale automatic vessel monitoring based on dual-polarization sentinel-1 and ais data," *Remote Sens.*, vol. 11, pp. 1078, 2019.

[24]. S. Wu *et al.*, "Deep high dynamic range imaging with large foreground motions," in *Proc. ECCV*, 2018, pp. 120–135.

[25]. S. Ioffe *et al.*, "Batch normalization: Accelerating deep network training by reducing internal covariate shift," in *Proc. ICML*, 2015, pp. 448–456.

[26]. Y. Zhang, Y. Tian, Y. Kong, B. Zhong, and Y. Fu, "Residual dense network for image super-resolution," in *Proc. CVPR*, 2018, pp. 2472–2481.

[27]. S. Woo *et al.*, "CBAM: Convolutional block attention module," in *Proc. ECCV*, 2018, pp. 3–19.

[28]. J. Hu *et al.*, "Squeeze-and-excitation networks," arXiv:1709.01507, 2017.

[29]. X. Wang *et al.*, "Non-local neural networks," arXiv:1711.07971, 2017.

[30]. K. He *et al.*, "Delving deep into rectifiers: surpassing human-level performance on ImageNet classification," in *Proc. ICCV*, 2015, pp. 1026–1034.

[31]. T. Zhang and X. Zhang, "Integrate traditional hand-crafted features into modern CNN-based models to further improve SAR ship classification accuracy," in *Proc. APSAR*, 2021, pp. 1–6.

[32]. J. He *et al.*, "Group bilinear CNNs for dual-polarized SAR ship classification," *IEEE Geosci. Remote Sens. Lett.*, vol. 19, pp. 1–5, 2022.

[33]. T. Zhang and X. Zhang, "Injection of traditional hand-crafted features into modern CNN-based models for SAR ship classification: What, why, where, and how," *Remote Sens.*, vol. 13, no. 11, pp. 2091, 2021.